big data and cognitive computing

*Article*

# Innovative Sentiment Analysis and Prediction of Stock Price Using FinBERT, GPT-4 and Logistic Regression: A Data-Driven Approach

Olamilekan Shobayo [1,*], Sidikat Adeyemi-Longe [1], Olusogo Popoola [1] and Bayode Ogunleye [2]

1. School of Computing and Digital Technologies, Sheffield Hallam University, Sheffield S1 2NU, UK; o.popoola@shu.ac.uk (O.P.)
2. Department of Computing & Mathematics, University of Brighton, Brighton BN2 4GJ, UK; b.ogunleye@brighton.ac.uk
* Correspondence: o.shobayo@shu.ac.uk

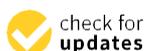

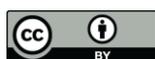

**Abstract:** This study explores the comparative performance of cutting-edge AI models, i.e., Finaance Bidirectional Encoder representations from Transsformers (FinBERT), Generatice Pre-trained Transformer GPT-4, and Logistic Regression, for sentiment analysis and stock index prediction using financial news and the NGX All-Share Index data label. By leveraging advanced natural language processing models like GPT-4 and FinBERT, alongside a traditional machine learning model, Logistic Regression, we aim to classify market sentiment, generate sentiment scores, and predict market price movements. This research highlights global AI advancements in stock markets, showcasing how state-of-the-art language models can contribute to understanding complex financial data. The models were assessed using metrics such as accuracy, precision, recall, F1 score, and ROC AUC. Results indicate that Logistic Regression outperformed the more computationally intensive FinBERT and predefined approach of versatile GPT-4, with an accuracy of 81.83% and a ROC AUC of 89.76%. The GPT-4 predefined approach exhibited a lower accuracy of 54.19% but demonstrated strong potential in handling complex data. FinBERT, while offering more sophisticated analysis, was resource-demanding and yielded a moderate performance. Hyperparameter optimization using Optuna and cross-validation techniques ensured the robustness of the models. This study highlights the strengths and limitations of the practical applications of AI approaches in stock market prediction and presents Logistic Regression as the most efficient model for this task, with FinBERT and GPT-4 representing emerging tools with potential for future exploration and innovation in AI-driven financial analytics.

## 1. Introduction

The prediction of stock market movements has been a focal point for researchers and investors due to the financial market's complexity and volatility. The ability to make precise stock or market predictions can result in improved decision-making, reduction of risks, and increased profitability. Traditional statistical techniques often fail to identify the complex patterns in stock data, especially when affected by external variables like news and market sentiment. Recent advancements in machine learning and deep learning have provided more sophisticated tools to address this problem. However, the primary research problem this study addresses is the need for more effective models that can effectively employ financial news sentiment to predict stock market trends. Sentiment analysis, a tool that analyzes the emotional tone or opinion behind financial news or social media, has become a crucial tool in stock market prediction, as investor sentiment significantly impacts market behaviors. The rise of Natural Language Processing (NLP) models, such as FinBERT and GPT-4, has created new opportunities for analyzing unstructured textual data like financial news [1]. These models can classify sentiment (e.g., positive, negative, neutral) and predict how such sentiment may influence stock prices. Moreover, simple and classic models like Logistic Regression remain effective for sentiment classification and market prediction when adjusted properly.

In this study, we aim to contribute to the intersection of finance and technology through empirical evaluation of the performance of advanced NLP models, i.e., FinBERT and GPT-4, as against traditional machine learning methods, such as Logistic Regression, for predicting the stock index trend. Each model was trained on historical NGX All Share Index data labels and financial news, utilizing various sentiment analysis techniques in the process. The labeled data allows the models to associate input news features with specific outcomes, which leads to better prediction accuracy when analyzing news articles for tasks like sentiment analysis or topic classification [2]. Five key metrics, i.e., Accuracy, Precision, Recall, F1 Score, and ROC AUC, were used to assess the performance of each model. Previous research has demonstrated the potential of FinBERT, a financial domain-specific model, in understanding financial terminology and context with remarkable precision [3]. However, its resource-intensive nature can present challenges in terms of computational efficiency. GPT-4, a versatile language model, has remarkable capabilities in understanding and generating human-like text. This makes it ideal for processing unstructured news data. Nevertheless, the exploration of its predetermined and heuristic approach may restrict its precision in particular financial situations. Logistic Regression is simple, computationally efficient, and produces reliable results when properly optimized. The results show that Logistic Regression outperformed both FinBERT and GPT-4 across most metrics despite their cutting-edge text analysis capabilities. Furthermore, this research contributes to the broad discussion of hybrid approaches that involve the integration of classic and advanced models to offer superior results. The findings suggest that Logistic Regression achieved the highest accuracy (81.83%) and ROC AUC (89.76%), while FinBERT and GPT-4 lagged behind in predictive accuracy. This highlights the effectiveness of traditional models when properly tuned and the future potential for hybrid approaches combining the strengths of NLP models with simple classic models for enhanced prediction accuracy. This study provides a framework for more effective and scalable market prediction solutions by bridging the gap between advanced NLP technology and conventional financial prediction techniques. This study not only assesses the predictive power of FinBERT, GPT-4, and Logistic Regression for stock market trends but also identifies the wider implications of artificial intelligence (AI) use in financial or equity markets. This suggests future directions for research to enhance financial forecasting through hybrid models.

*1.1. Study Hypothesis*

**Hypothesis 1 (H1).** *Machine learning models, such as FinBERT, GPT-4, and Logistic Regression, can effectively classify financial news sentiment.*

**Hypothesis 2 (H2).** *Domain-specific models like FinBERT and GPT-4, with their generalized and powerful natural language understanding, will outperform classic models like Logistic Regression in accurately capturing market sentiment due to their ability to better handle context, financial jargon, and nuanced market sentiment expressions.*

**Hypothesis 3 (H3).** *GPT-4, a general-purpose language model, can achieve high accuracy in sentiment analysis but may still underperform compared to a fine-tuned and domain-specific model like FinBERT for tasks involving specialized financial terminology.*

*1.2. Practical Significance of the Study*

The practical significance of this study stems from its capacity to support various stakeholders in the stock market or financial industry. Organizations can make more informed decisions by using machine learning models to analyze market sentiment in financial news. In particular, the study provides a framework for:

- Financial Analysts: It assists financial analysts in the automation of sentiment analysis of financial news that moves the stock market, thereby improving decision-making in real time.
- Investors: It assists investors with strategic investment decisions by providing insights into market trends based on the tone of pertinent market news.
- News Platforms: It improves news platforms' capacity to prioritize and filter content according to market sentiment, which offers more value to the users.
- Data Scientists: It offers data scientists a comparison of different machine learning models, which helps in the selection of appropriate models for financial sentiment analysis tasks.
- Organizations and AI Researchers: It is valuable for companies and artificial intelligence researchers who are interested in comparing the performance of domain-specific models with more general models across different financial data tasks.

To provide a clear understanding of the current state of research and position this study within the broad academic context, a comprehensive literature review was developed below. This entails the different data analysis methods employed in previous stock price prediction studies, along with their respective strengths and weaknesses. Furthermore, the performance of each model will be compared with existing approaches to demonstrate their ability to gauge the market's mood and predict stock price movements. Positive sentiment indicates rising stock prices, while negative sentiment signals potential declines, which makes it a useful tool in stock market prediction and decision-making.

*1.3. Literature Review*

Liu et al. developed a pre-trained FinBERT model for financial text mining and sentiment analysis. Their research demonstrated that FinBERT significantly outperformed other models in understanding financial language, providing more accurate sentiment classifications in financial reports and news [3]. However, the study was limited by its focus on FinBERT only without evaluating the performance of other models. This study builds upon their findings by applying and evaluating FinBERT with other AI models to ascertain its utility and performance on financial news text.

Leippold explored the vulnerabilities of financial sentiment models to adversarial attacks on GPT-3. The research revealed that subtle manipulations in financial texts could alter sentiment predictions, which highlight GPT-3's sensitivity to adversarial inputs. Although GPT-3 showed great potential in financial text generation, its lack of interpretability remains a significant limitation [4]. The research extends this work by integrating explainable AI methods alongside GPT-4 to improve transparency in financial sentiment and predictions. This offers a more robust approach to understanding model decision-making.

Yang et al. combined LASSO, LSTM, and FinBERT to predict stock price direction using technical indicators and sentiment analysis [5]. Their model achieved high accuracy in predicting price movements based on market sentiment. However, their approach was limited by the feature extraction techniques that were employed, which may not fully capture non-linear relationships in financial data. This study addresses feature extraction issues by strictly extracting financial news and adding NGX labels for easy topic classification as part of input features.

Sidogi et al. used FinBERT and LSTM to analyze the impact of financial sentiment on stock prices. Their study focused on using LSTM for time series forecasting, with FinBERT providing sentiment features from financial news and reports. The study limited performance metrics to only root mean square error (RMSE) and mean absolute error (MAE) without evaluating the performance of FinBERT itself [6]. This research intends to evaluate all selected models to ascertain each model's performance for better perspectives.

## 2. Materials and Methods

This study uses news headlines sentiment data scraped from Nairametric and Proshare websites using Data Miner to analyze the performance and direction of the Nigerian stock market. Nairametric is a Nigerian investment advocacy company, while Proshare is a professional practice firm that offers various services to connect investors and markets. The aim of using two organizations' news was to ensure accuracy, reduce errors, and boost trust. News headlines were chosen over social media for sentiment analysis due to their credibility, structured data, reduced bias, event-centricity, manageable volume, less manipulation, reliability, timeliness, source verification, journalistic standards, and focused content. The scraped news data included major market, financial economic, and listed companies' news. The news data spanned from 4 January 2010 to 7 June 2024, comprising a total of 24,923 news headlines. These news headlines were aggregated into 3573 distinct temporal observations, providing a detailed dataset for this time frame. News labels are based on the stock index categorization, where:

"Class 1" implies daily share price gain, and "Class 0" signifies unchanged or fall in share price.

### 2.1. News Data Preprocessing and Preparation

The scraped news headlines were preprocessed using the Natural Language Toolkit (NLTK). The NLTK is a Python library used for dataset cleansing and natural language processing (NLP). The data cleaning methods include stopword elimination, data conversion, concatenation, tokenization, noise abatement, normalization, and feature extraction.

- Stopwords: High-frequency words with limited semantic meaning (e.g., "the", "is", etc.) were removed to improve model accuracy. This enables focus on more meaningful terms.
- Data conversion: All text was converted to lowercase to avoid treating the same word differently based on capitalization.
- Concatenation: Text strings were combined where necessary to ensure the dataset was well organized for feature engineering and financial analysis.
- Tokenization: Text was split into tokens (manageable units) to make it easier for the models to process. This allows for better handling of the data.
- Noise abatement: Unnecessary characters, symbols, or data that mask market trends and reduce data analysis were removed to enhance the clarity of the market trends and improve the precision of the sentiment analysis.
- Normalization: This is a process that standardizes text to improve the speed and quality of text analysis. Stemming and lemmatization are methods used to standardize words by removing suffixes and affixes to reveal their root form. Stemming algorithms use heuristic principles for efficiency and simplicity. Heuristic principles use pattern matching, rule-based simplifications, fixed-order operations, search space reduction, and statistical heuristics to guide problem-solving and decision-making [7]. Lemmatization analyzes the context and grammatical components to generate a lemma (root word), which improves text analysis, accuracy, and clarity through contextual comprehension [8]. These allowed the models to better interpret and analyze the meaning of the text across different contexts.
- Feature extraction is a method that transforms data into features for machine and deep learning algorithms [9]. It improves data interpretability, model performance, and dimensionality. The model-specific text feature extraction or news text preparation includes:
    (a) FinBERT: BERT embedding, a state-of-the-art technique that captures the context of words in a sentence, was employed. This method allowed the model to understand the meaning behind specialized financial terminology and subtle expressions, which is crucial in sentiment analysis. BERT's transfer learning potential also made it particularly resilient in tasks like sentiment classification and identifying false news.

(b)　GPT-4: GPT-4, as a large language model, does not require explicit feature extraction techniques like TF-IDF or BERT embeddings. It leverages its pre-trained architecture to understand and generate responses based on the input text. The main advantage of GPT-4 is that it is already equipped with knowledge from a wide array of domains and, as a result, requires minimal data preprocessing. However, some preprocessing steps, such as stopword removal, lowercase conversion, concatenation, tokenization, noise abatement, and normalization, were still applied to ensure consistency in the data before passing it to GPT-4.

(c)　Logistic Regression: TF-IDF (Term Frequency–Inverse Document Frequency) vectorization was used to transform the cleaned text data into features. TF-IDF is well suited for sparse and high-dimensional datasets like financial news, as it captures the importance of words in individual documents relative to the overall dataset. This method helped the model focus on important words while efficiently managing large amounts of data.

The text data preparation methods used were tailored to the specific strengths of each model. For Logistic Regression, TF-IDF was selected for its efficiency and interpretability, while BERT embeddings were used for their ability to capture context and handle the nuances of financial news sentiment. The careful preprocessing of the news data and selecting the appropriate methods for each model assisted in addressing the unique challenges posed by financial sentiment analysis.

*2.2. Data Preparation*

The preprocessed news headlines datasets were split into 70% for training, 15% for validation, and 15% for testing. The dataset is a chronological dataset, and its temporal order was maintained throughout the model training and evaluation. This makes traditional cross-validation, such as k-fold cross-validation, not appropriate for this study since it randomly shuffles the data. This activity could lead to the leakage of future information into the training set. Time series cross-validation (TSCV) was used to train and validate the model with folds ($n$ = 5) on the news dataset. This method maintains the time dependency between data points. TSCV closely mirrors how the model would perform in real-world applications, such as stock price prediction, by simulating real-world scenarios with unseen future data in each fold. This approach enhances the model's ability to generalize, reduces overfitting, and improves prediction accuracy [10]. Moreover, it avoids data leakage by ensuring that no future information influences the training phase, which results in a more reliable evaluation of the model's performance. This method uses all available data points across the folds for both training and validation, providing a comprehensive assessment of the model's robustness and accuracy. Additionally, data labels based on the stock index categorization were added to the news dataset as an input feature of the model. The labels help the model to differentiate news categories, reduce noise, and enable efficient model training through clear mappings between input features and the desired outcomes [11].

*2.3. Algorithm Selection and Computation for Financial News*

The study used FinBERT, GPT-4, and Logistic Regression models to find the optimal method because of their ability to handle complex language and domain-specific text and provide interpretable results. Logistic Regression (LR) was selected for this study, among other machine learning models, because of its simplicity, interpretability, and effectiveness in binary classification tasks and alignment with NGX stock index label categorization. Additionally, LR was also considered because of its ease of interpretation, computational efficiency, and baseline comparison in text classification [12]. This provides a point of comparison for more complex models like GPT-4 and FinBERT. Although other machine learning methods, such as support vector machines (SVMs) or random forests, could also provide good performance, LR is a tried-and-true technique for text-based sentiment analysis. Its robustness and simplicity make it a good choice when interpretability is a priority. GPT-4 is a general-purpose model that is not specifically tailored for financial news. However, it was considered for this study due to its versatility and capacity to handle a wide range of text analysis tasks. Furthermore, domain-specific models like FinBERT were used due to the empirical evidence of offering higher accuracy in capturing nuanced sentiment in financial reports due to their training on specialized financial data [13]. Other machine learning methods could be explored in the future for further accuracy improvement. The employment of the three AI models enables a comprehensive assessment of distinct algorithmic methodologies, each possessing distinctive advantages. This methodology allows researchers to comprehend the compromises between the intricacy of the model, its performance, and its interpretability. It also empowers us to make well-informed choices regarding the most suitable algorithm for the particular use case of the AI model.

*2.4. FinBERT Architecture, Development, and Training*

FinBERT, a specialized variant of BERT, is a pre-trained language model designed for financial text analysis that interprets and analyzes the nuances of financial language, including finance- and economics-specific jargon. It excels in financial sentiment analysis, market sentiment research, stock trading strategy formulation, and risk management. FinBERT-base model (12 layers, 768 hidden size, 12 attention heads) and FinBERT-Large (24 layers, 1024 hidden size, 16 attention heads) are the variants, but the study used FinBERT-base because of its computational efficiency, lower memory requirement, sufficient performance, overfitting concerns, and ease of use. The FinBERT-base model training process started with ascertaining the data integrity through correct parsing of the date fields of the cleaned news data to prevent errors in temporal analyses. The Pandas library was used for its robustness in handling large datasets efficiently. The cleaned data were tokenized using input IDs and attention masks, setting a maximum sequence length of 128 to handle long texts efficiently, and converted into PyTorch tensors for data feeding into the model. This assisted in debugging and monitoring the process. Dataloaders were created for batching data during training, validation, and testing. Huang et al. reported on FinBERT as a model for extracting information from financial text and emphasized the importance of domain-specific tokenizers for enhancing the model performance in finance fields [12]. News data were divided into training, validation, and test sets in a chronological manner to maintain the temporal order of the new financial data. This was performed to prevent data leakage and ensure the model was tested on unseen future data. Hyperparameter tuning was performed using Optuna, while automatic mixed precision (AMP) training was employed to fast-track the training process while maintaining model accuracy. Optuna recommended the optimal learning rates and batch sizes to maximize validation performance. Von der Mosel et al. conducted a study on BERT transformer models that were trained with software engineering data and general domain models. The study highlighted the usefulness of AMP in enabling faster computation, reducing memory usage on GPU, and allowing for larger batch sizes without increasing hardware requirements [14]. Also, early stopping with a patience parameter of 5 was applied to prevent overfitting and preserve the model's generalization capabilities. The final trained model was evaluated with classification metrics on the test dataset.

*2.5. GPT-4*

GPT-4 is a general-purpose model, meaning it was not specifically fine-tuned for financial news but has broad language understanding capabilities. The findings reveal that GPT-4 can perform sentiment analysis and classification using classic machine learning models such as predefined approach, Naïve Bayes, linear regression, etc. However, this study explores the predefined sentiment approach of GPT to classify, analyze, and generate sentiment scores on financial news data. The process started with uploading the minimal preprocessed financial news headline csv file with the instruction of using GPT to split into 70% training, 15% validation, and 15% testing (according to time frame) and to classify, evaluate, and perform sentiment analysis on the uploaded data. The preprocessed news data were prepared and maintained in their temporal order format, ensuring compatibility with the GPT-4 Application Programming Interface (API). The preprocessed news headlines were then fed into the GPT-4 API. The passing of data through the API enables GPT-4 to process each headline internally using its predefined approach to analyzing the context, semantics, and sentiment of the news articles. GPT-4 employs a combination of natural language understanding and pattern recognition to assess the sentiment and classify each news item. This method demonstrates the efficiency of end-to-end processing of GPT as it analyzed news data from preprocessed headlines to internal representations (such as sending news data to GPT-4 API; GPT-4 processes text and analyzes sentiment of news text) and ended with sentiment scores and classification output for evaluation. Performance metrics like accuracy, precision, recall, and ROC AUC are used to evaluate GPT-4's effectiveness in financial sentiment analysis. This method does not require manual feature extraction and is capable of handling complex language. However, it may not always capture domain-specific financial nuances like a model fine-tuned for this purpose.

*2.6. Logistic Regression Architecture Development and Training*

Logistic regression is a statistical model that is designed to solve binary classification problems. The architecture of the Logistic Regression (LR) used for this study is defined by its key parameter components and hyperparameters. LR binary classification fits into listed stock labels, with "Class 1" representing daily share price gain and "Class 0" signifying price decline or unchanged price. This assisted in predicting the financial news input's class. The model architecture core is the 'C' parameter, penalty, and solver. The value of C balances fitting training data well and keeps the model simple to minimize overfitting. A smaller C discourages large coefficients, strengthening regularization, whereas a bigger C weakens it. Also, L2 regularization (penalty = 'l2') kept model coefficients minimal to prevent overfitting. 'Liblinear' was chosen over 'lbfgs' and 'saga' due to its resilience and speed. Liblinear optimizes small datasets quickly and reliably. It also enables rapid computation for L2 penalized Logistic Regression. The sigmoid function, another key Logistic Regression component, translates all real numbers to values between 0–1. It used scikit-learn to create a sentiment score and 0.5 as a decision threshold to classify inputs. This is based on logistic function probability, and the mathematical formula for Sigmoid is

$$Sigmoid(z) = \frac{1}{1 + e^{-z}} \quad (1)$$

where $z$ = the weighted sum of the input features; $e$ = mathematical constant (~2.71828); and $-z$ = negative of the input $z$.

The LR model was trained using Optuna to automatically optimize "C" and solver hyperparameters. The objective function was defined, and parameter ranges were suggested to maximize the F1 score: C: 0.0001–100; Solver: Liblinear, lbfgs; and Penalty: L2. Optuna explored multiple C values and tested various solvers to identify the optimal combination that yields the best F1 score on the validation data. Additionally, time series cross-validation with $n = 5$ was implemented to stabilize the selected hyperparameters across several time-based folds. The model was trained on the training dataset and evaluated on the validation dataset to calculate the F1 score. F1 was selected over other classification metrics because of its suitability in scenarios with class imbalance. This process was repeated for each set of hyperparameters suggested by Optuna. The optimal hyperparameters were chosen, and the LR model was retrained on the training and validation sets and tested on a news testing set. These methods provide a reliable method for developing, optimizing, and training this study's LR model.

*2.7. Hardware and Computational Resources*

A premium NVIDIA A100 GPU (Graphics Processing Unit) option on Google Colab, cloud-based computational resources, was employed. This allows for the efficient execution of the computationally intensive FinBERT model. It also offers high memory capacity, reduces the processing time from the initial days to minutes, and makes the task feasible. FinBERT's total processing time is around 110 min, with training time (fine-tuning time) of 90 min and testing and inference time of 10 min. The resource consumption includes the A100 GPU's 40 GB of VRAM, which is crucial for managing the memory requirements of FinBERT. This VRAM stores the model's data, processes, and intermediate computations, allowing FinBERT to run efficiently without memory limitations. Also, the CPU is used for tasks like data loading, text tokenization, input/output operations, and data batch preparation for the GPU. The CPU tasks are essential for the overall process, and the usage is moderate and not as computationally intensive as the deep learning computations on the GPU. Therefore, FinBERT does require more computation power (including higher GPU and processing capabilities) and is less time efficient with the training and inference times, which were longer than for Logistic Regression. This could be attributed to its transformer (deep learning) architecture and explains why the local laptop was unable to handle FinBERT demands. The available laptop hardware resources could run the Logistic Regression due to its simplicity.

## 3. Results

### 3.1. FinBERT

The optimal hyperparameter suggested by Optuna was used to train the FinBERT model on data in chronological order and tested on a 15% test dataset. The evaluation results are shown in Table 1.

**Table 1.** Evaluation results of FinBERT.

| Best Hyperparameters: Learning Rate: $3.564937469182303 \times 10^{-5}$, Batch Size: 16 | | |
|---|---|---|
| **Test Set Metrics** | | **Test Metrics (Percentage)** |
| Accuracy | 0.6333 | 63.33 |
| Precision | 0.6376 | 63.76 |
| Test Recall | 0.6333 | 63.33 |
| Test F1 Score | 0.6330 | 63.30 |
| Test ROC AUC | 0.6559 | 65.59 |

#### 3.1.1. Model Evaluation

The result of the evaluation metric above shows that FinBERT correctly predicts the sentiment of financial news 63.33% of the time. This shows moderate performance. The complexity and fluctuating nature of financial terms and market sentiments can make it tough to achieve a high level of accuracy. The precision score shows that when FinBERT predicts the news sentiment, it is correct 63.76% of the time. This performance is also modest. The prediction precision is very important, as false positive sentiment can lead to incorrect market trading decisions. The recall result shows FinBERT has been able to identify 63.33% of all relevant instances of sentiment. This implies the model is moderately efficient, as it might still miss some relevant market signals and sentiments in the financial data. The F1 score is a balance between precision and recall, and the score at 63.30% is fairly balanced but not strong enough. The ROC AUC score of 65.59% is a positive indicator of the model's ability to differentiate between positive and negative sentiment. An ROC AUC value close to or higher than 0.7 is considered good in a complex field like financial news interpretation, as it indicates a better ability to distinguish between positive and negative sentiment. Chen et al. empirically demonstrated benchmarking scores of existing methods and discussed specific models designed for financial news sentiment analysis [15]. The findings suggested that well-performing models in complex financial sentiment analysis often achieve ROC AUC scores close to 0.7 or higher. This confirms the research claim that FinBERT does moderately well in predicting financial news. Overall, the FinBERT prediction performance is within range, given the complexity of financial news and the fact that sentiments are hidden in technical language and influenced by context. The studies of Kirtac and Varghese emphasize that while models like FinBERT can perform well on financial data, they often require specific adaptations to the financial datasets they analyze [16,17]. The FinBERT scores indicate an effective but not highly reliable model for critical financial decisions.

3.1.2. Visual Inspection

Figure 1 shows the ROC AUC value of 0.66 which implies that the FinBERT's accuracy is modest. This demonstrates that the FinBERT model is reasonably capable of distinguishing between positive and negative classes, but there is room for improvement.

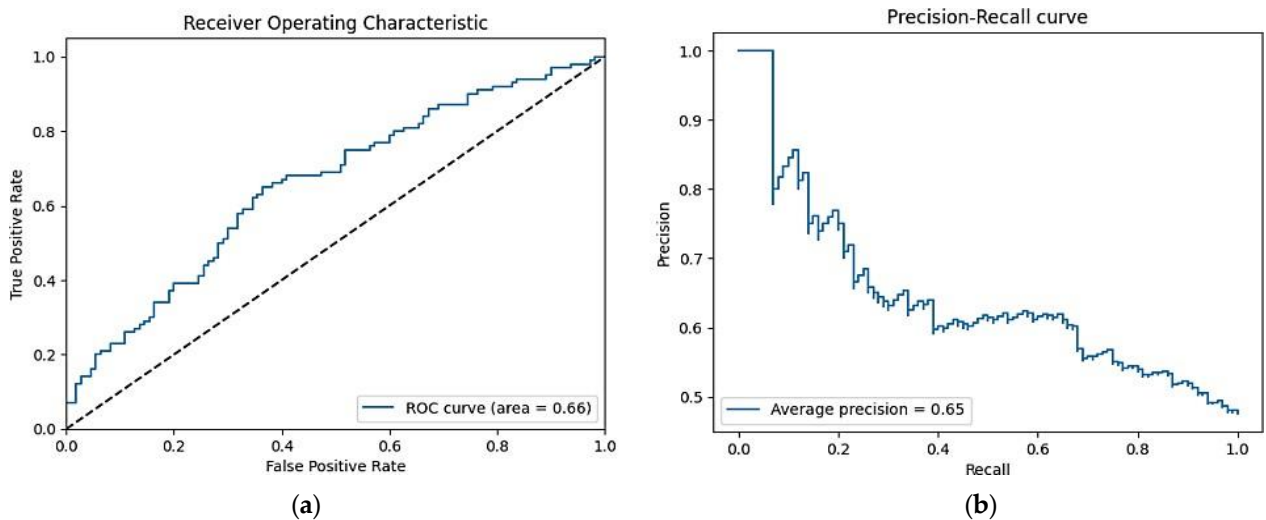

**Figure 1.** (**a**) ROC curve and (**b**) precision–recall curve of FinBERT model.

The precision versus recall curve also has an average precision score of 0.65. This indicates moderate performance and also shows that the model struggles with maintaining high precision as it tries to increase recall. The curve suggests that the model is effective in specific scenarios (low recall) but struggles to balance both high precision and high recall simultaneously.

The daily sentiment score plot (Figure 2a) exhibits frequent daily fluctuations in sentiment, showing a high level of variability. This suggests that the underlying news or events that drive sentiment are changing frequently and impacting the overall market or public sentiment on a daily basis. This volatility may reflect a highly reactive market environment that shifts investor opinions and can contribute to sharp, short-term movements in stock prices. There is no long, sustained period where the sentiment remains at the upper or lower bounds. There are periods where sentiment appears to cluster around certain ranges. For example, around 2021 and 2023, there seems to be more clustering of sentiment in the middle range. This could indicate a more neutral sentiment during those years. Other periods (like 2010–2012 and 2020–2022) seem to have more extreme higher sentiment scores near 0.8 to 0.9, suggesting periods of particularly strong positive sentiment despite volatility. The data continue to show high variability from 2023 to 2024, with scores ranging across the entire spectrum. This suggests that market or public sentiment has remained volatile, without strong trends toward sustained positivity or negativity. The high volatility and constant shifts in sentiment suggest that investors should be prepared for rapid market changes. Traders might use short-term strategies, such as momentum trading, to capitalize on these frequent shifts in sentiment.

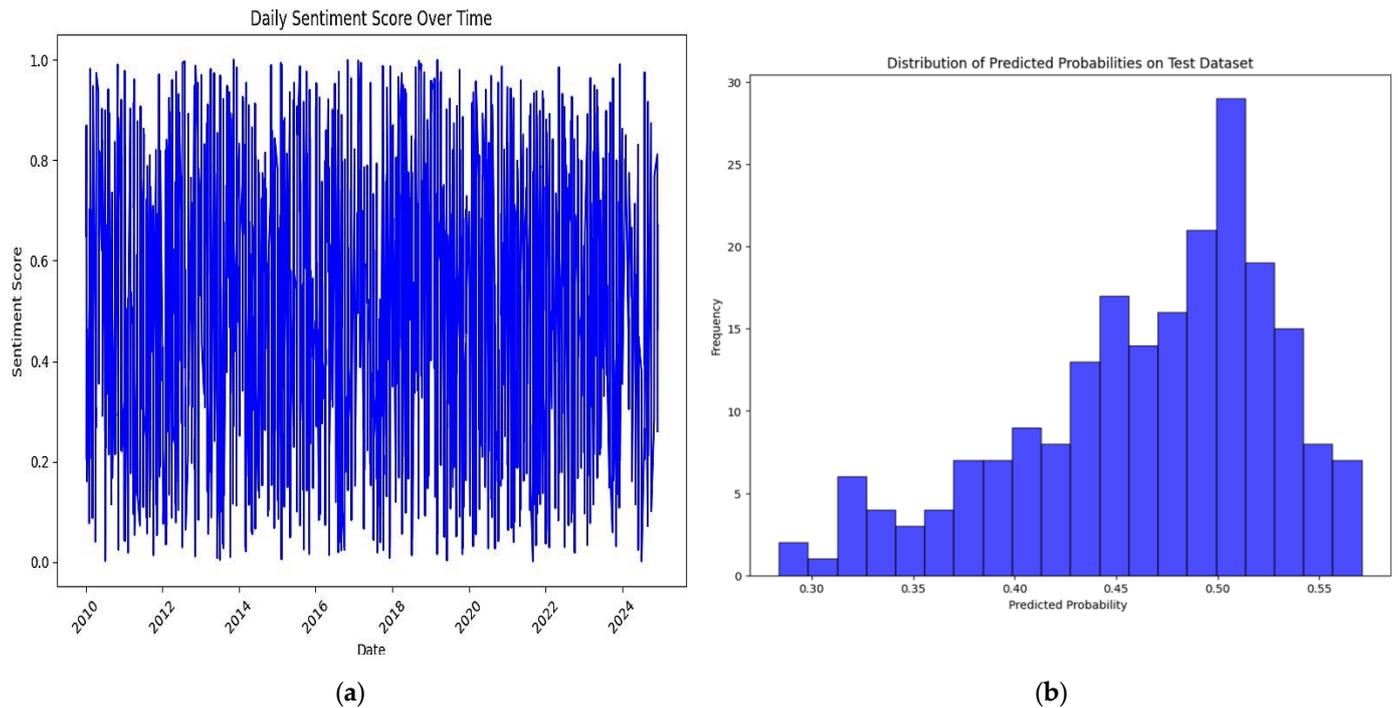

**Figure 2.** (**a**) Sentiment score over time and (**b**) distribution of predicted probabilities of FinBERT model.

　　The distribution of predicted probabilities on test data (Figure 2b) shows that the majority of the predicted probabilities are centered around 0.5, which suggests that the model often predicts a roughly equal probability of either stock price increase or decrease. This shows that the FinBERT model is likely uncertain in most of its predictions. The predictions hovering around 0.5 indicate that the model does not strongly favor one outcome over the other. This might suggest a market in balance or one with mixed signals. The prediction range from 0.30 to 0.55 shows conservative predictions without high confidence on extreme probabilities close to 0 or 1. The noticeable peak around 0.50 indicates the model sees equal chances of either outcome; the market might not have a clear direction or might be a signal of an uncertain market. There is room for improvement in the model confidence level.

　　Overall, the FinBERT model shows moderate effectiveness in classification, with some uncertainty in predictions. The sentiment score fluctuations provide insights into market sentiment over time, which can be valuable for trading decisions. Both the ROC and precision–recall curves suggest that while the model performs above random chance, there is potential for further refinement to improve its accuracy and reliability.

*3.2. GPT*

　　The response from GPT-4 itemized the steps of executing the instruction as data loading; text inspecting; text data preprocessing using "re" library for a predefined approach; splitting the data into 70% training, 15% validation, and 15% testing; performing sentiment analysis; classification; and evaluating the model on the validation and test sets. The predefined approach based the evaluation on predefined sentiment labels. Figure 3 below shows the financial news inputted into GPT-4:

**Financial Data**

|   | Date       | label | text                                                                            |
|---|------------|-------|---------------------------------------------------------------------------------|
| 1 | 07/06/2024 | 1     | world bank mandates hire security consultant billion loan project nigeria despite |
| 2 | 06/06/2024 | 0     | afrexim bank disburse another million nnpc billion loan n trillion provision fuel |
| 3 | 05/06/2024 | 1     | executive order tinubu stop …                                                    |

**Figure 3.** The financial news data input into GPT.

3.2.1. Model Evaluation

Table 2 shows the results of the evaluation metrics using GPT-4.

**Table 2.** Evaluation metric results of GPT-4.

| Evaluation Metrics   | Predefined Sentiment Set Metrics | Predefined Sentiment Set Metrics (Percentage) |
|----------------------|----------------------------------|-----------------------------------------------|
| Validation Accuracy  | 0.5720                           | 57.20                                         |
| **Test Evaluation**  |                                  |                                               |
| Accuracy             | 0.5419                           | 54.19                                         |
| Precision            | 0.7266                           | 72.64                                         |
| F1 Score             | 0.4509                           | 45.09                                         |
| Recall (Sensitivity):| 0.3269                           | 32.69                                         |
| AUC-ROC              | 0.6537                           | 65.37                                         |

The accuracy of 57.20% on the validation set indicates that the model correctly classifies sentiment more than half of the time. This shows moderate performance and suggests room for improvement in capturing the nuances of sentiment in financial news. The test accuracy of 54.19% is slightly lower than the validation accuracy. This suggests that the model may struggle to generalize on unseen data. This could be due to the inherent complexity of financial sentiment. The test precision of 72.66% is relatively high. This shows that when the model predicts a positive sentiment, it is often correct. The news with positive sentiment that has been selected is expected to genuinely reflect confidence about the state of the stock market. The low test recall of 32.69% suggests that the model misses many actual positive sentiment news. This indicates that many potentially significant positive news items might not be recognized. This leads to the underrepresentation of positive sentiment. The F1 Score of 45.09% is a balanced measure that shows the overall performance. It demonstrates the difference resulting from increased accuracy and decreased completeness. The Area Under the Curve—Receiver Operating Characteristic (AUC-ROC) of 65.37% is moderate and it suggests the model has a fair ability to differentiate between positive and negative sentiments. This is important for predicting the impact of news on stock movements. The predefined approach of GPT can reliably identify positive sentiments when they occur. This can be useful for stockbrokers or investors who are focusing on signals for bullish market conditions. However, the low recall means many positive opportunities might be missed. This potentially may lead to conservative trading strategies. The moderate AUC-ROC shows the predefined approach can capture some trends, but many might go unnoticed. This affects market prediction accuracy. Overall, the model performance suggests it should not be solely relied upon for stock trading decisions.

3.2.2. Visual Inspection

The ROC curve shown in Figure 4, with an AUC of 65.37%, shows a moderate ability to differentiate between positive and negative market sentiments. However, it cannot be compared to the precision–recall curve, which fluctuates significantly in middle recall values. The precision–recall curve below starts high but drops as recall increases, indicating a trade-off between capturing more positive sentiments at the expense of accuracy. Senapaty et al. emphasize the importance of considering both ROC and precision–recall curves to understand the trade-offs between sensitivity, specificity, and precision in practical applications [18].

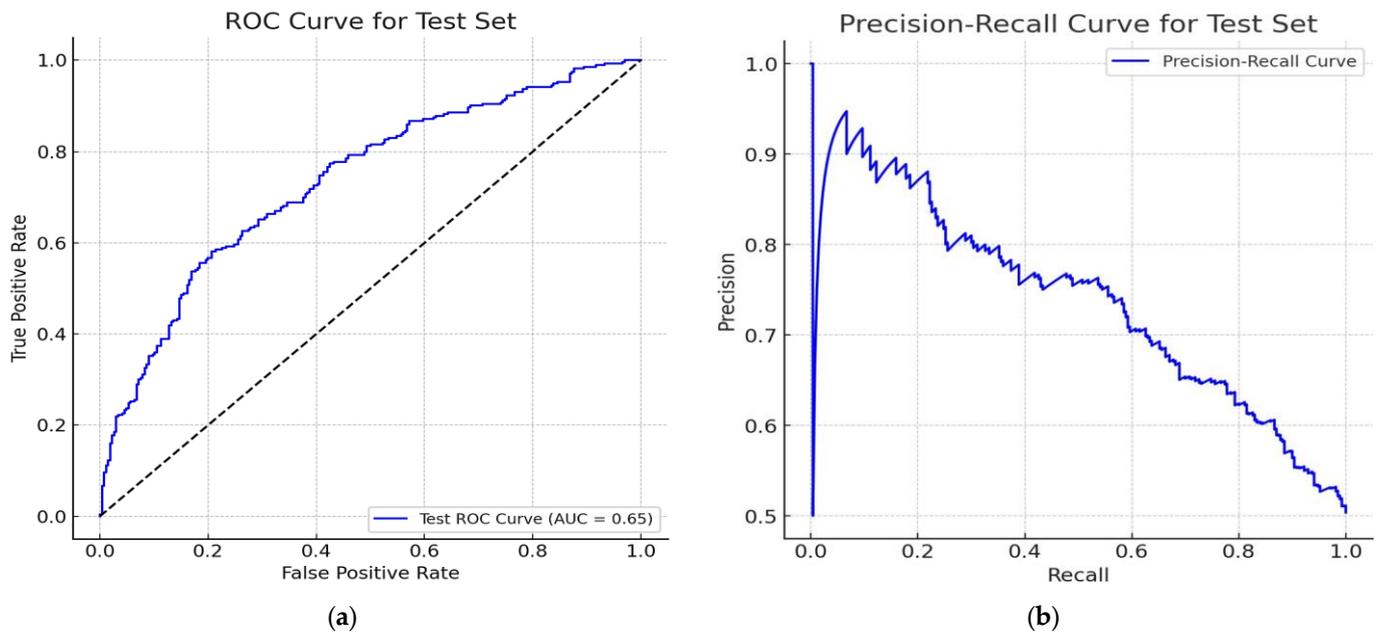

**Figure 4.** (**a**) ROC curve and (**b**) precision–recall curve of GPT.

The GPT model predicts significant daily fluctuation in sentiment throughout the entire time period as shown in Figure 5a,b. The wide range of sentiment scores suggests that the underlying events, market conditions, or financial news driving these scores are highly dynamic. The sentiment scores around 2012–2014 and 2017–2018 seem to trend high and cluster around 0.6 to 0.8. This could indicate more favorable news or market conditions during these periods that lead to more positive sentiment. The sentiment scores appear to be lower, on average, from 2022 onward, with many scores closer to the 0.4 to 0.6 range. This suggests more neutral or slightly negative sentiment in recent times. It also depicts less optimistic market news during the period. This pattern provides insights into market sentiment trends and potential impacts on stock prices. The constant sentiment fluctuation in response to market events reflects how the stock market and public perception are impacted by frequent changes in financial and economic news and conditions.

The predicted probabilities are clustered around 0.5 thresholds, with a central tendency toward the 0.45 to 0.50 range. The model's low confidence level indicates uncertainty in classifying financial news as positive (close to 1) or negative (close to 0) sentiment in numerous instances. The model makes relatively fewer predictions with probabilities below 0.30 or above 0.60. This suggests that the model is hesitant to assign extreme probabilities (either close to 0 or 1. This shows rare predictions with high certainty. This could be a result of a complex dataset where clear patterns are hard to detect and, as a result, cause the model to hedge its predictions toward the middle. The clear peak at around 0.50 suggests that the model is effectively predicting a "coin flip" scenario (i.e., where positive or negative are equally likely outcomes) for a significant portion of the test set. The roughly bell-shaped histogram suggests the model tends to predict outcomes with moderate probabilities most often and is less likely to predict with high confidence.

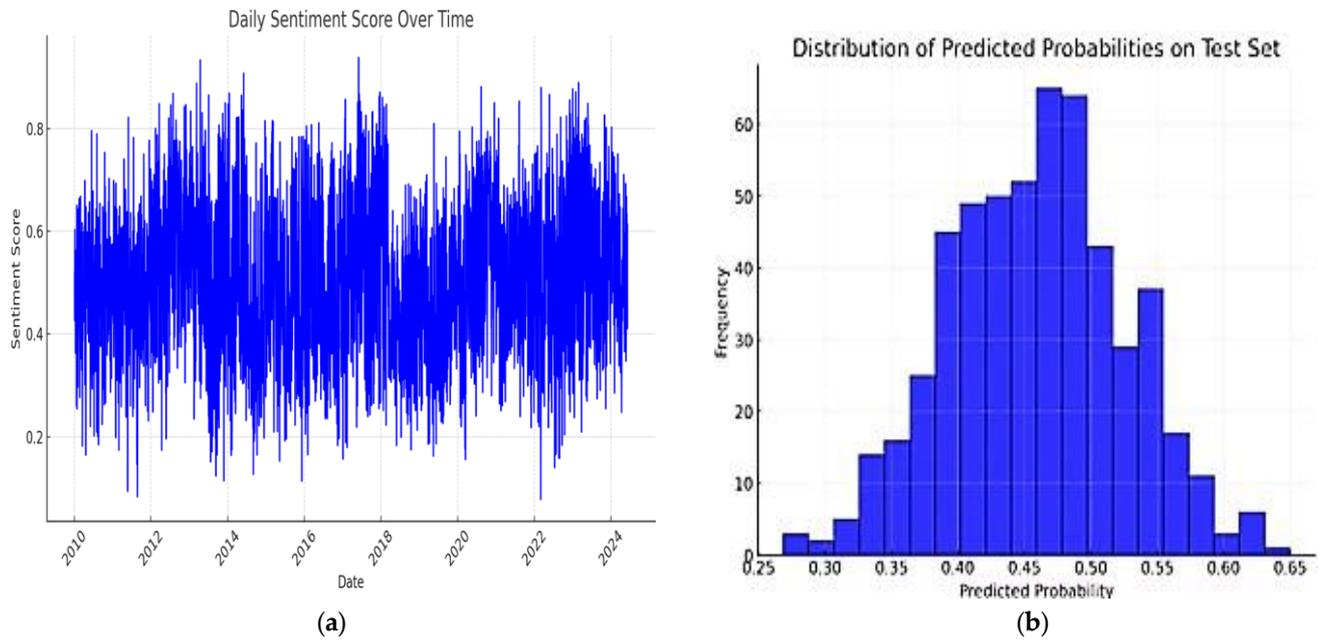

(**a**) (**b**)

**Figure 5.** (**a**) Sentiment score over time and (**b**) distribution of predicted probabilities of GPT.

*3.3. Logistic Regression*

The study used Logistic Regression to analyze sentiment in financial news headlines. The model was optimized using a logarithmic scale with a fixed penalty type of 'l2'. The model outputs were used to derive sentiment scores, which indicate the positive or negative sentiment expressed in each headline. The optimal hyperparameters from Optuna were used for model training and testing. The daily sentiment score was generated, and the evaluation results are shown in Table 3.

**Table 3.** Evaluation results of Logistic Regression.

| Best Hyperparameters: {'C': 3.037005064126959, 'Solver': 'Liblinear', 'Penalty': 'l2'} Training Metrics Accuracy: 0.8093 = 80.93% | | |
|---|---|---|
| **Test set metrics** | | Metric% |
| Accuracy | 0.8183 | 81.83 |
| Precision | 0.8257 | 82.57 |
| Test Recall | 0.8115 | 81.15 |
| Test F1 Score | 0.8185 | 81.85 |
| Test ROC AUC | 0.8976 | 89.76 |

3.3.1. Model Evaluation

The test results from using Logistic Regression on news data show an accuracy of 81.83%. This clearly shows a solid performance in distinguishing between positive and negative sentiments in financial news. Additionally, the model test accuracy is very close to the training accuracy of 80.93%, which indicates that the model generalizes well and there is no significant overfitting. The high precision of 82.57% implies that positive news is well identified, and out of all the positive predictions made by the model, 82.57% were actually positive. This shows that the model is quite good when it predicts positive sentiment and has a relatively low rate of false positives. This is very important for investors who rely on positive market signals to make buying decisions. The recall of 81.15% shows that out of all the actual positive cases, 81.15% were correctly identified by the model. This shows the model is doing well at capturing most of the actual positives. However, a slight drop in precision indicates a small trade-off between precision and recall. For stock market applications, this suggests potential for investment opportunities. The ROC AUC score of 89.76% shows the model's ability to distinguish between the positive and negative classes. A score closer to 100% is ideal. The results show excellent performance and indicate that the model is very good at ranking positive cases higher than negative cases. This is especially useful in scenarios where you might want to adjust the decision threshold for different costs of false positives and false negatives. The F1 Score of 81.85% is the harmonic mean of precision and recall that balances the two metrics. A value of 81.85% shows a good balance between precision and recall. This suggests a well-balanced model that is effectively managing both false positives and false negatives. The model generalizes well. This is shown by the close alignment of training and test accuracy. Additionally, there is a good balance between precision and recall. This means that the model is well suited for stock market sentiment where both false positives and false negatives are costly. Overall, the model is robust in distinguishing between the classes; this makes it reliable for the study.

3.3.2. Visual Inspection

The ROC curve and precision–recall curve in Figure 6a,b shows the performance of the Logistic Regression model on the financial news.

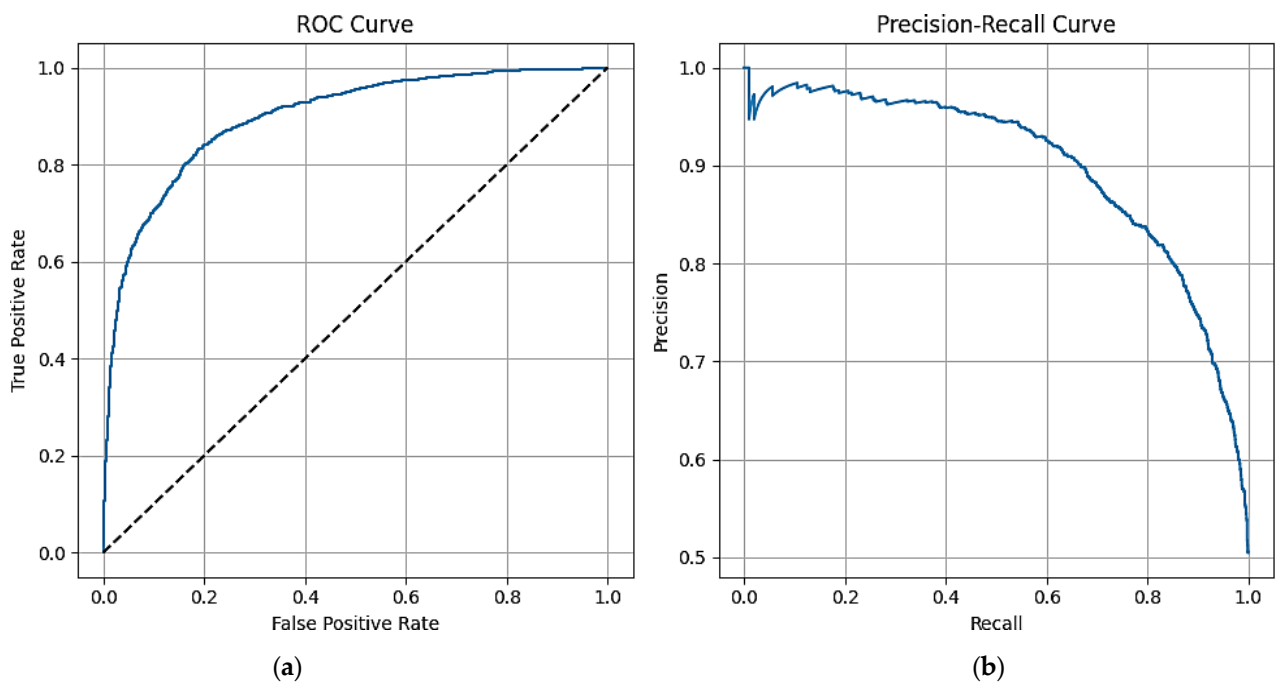

**Figure 6.** (**a**) ROC curve and (**b**) precision–recall curve of Logistic Regression model.

The ROC curve above the diagonal indicates superior performance compared to random guessing. The elevated AUC value indicates robust model performance. The model maximizes the true-positive rate and maintains a low false-positive rate. Logistic Regression effectively separates between positive and negative classes, as observed in the ROC curve surpassing the diagonal line. The precision–recall curve shows a positive correlation between precision and recall, with a decrease in precision as recall increases. Nevertheless, the model maintains high accuracy across different recall values, which indicates its performance across different thresholds. Furthermore, the Figure 7 shows the daily sentiment score over time and the distribution of predicted probabilities on the test set.

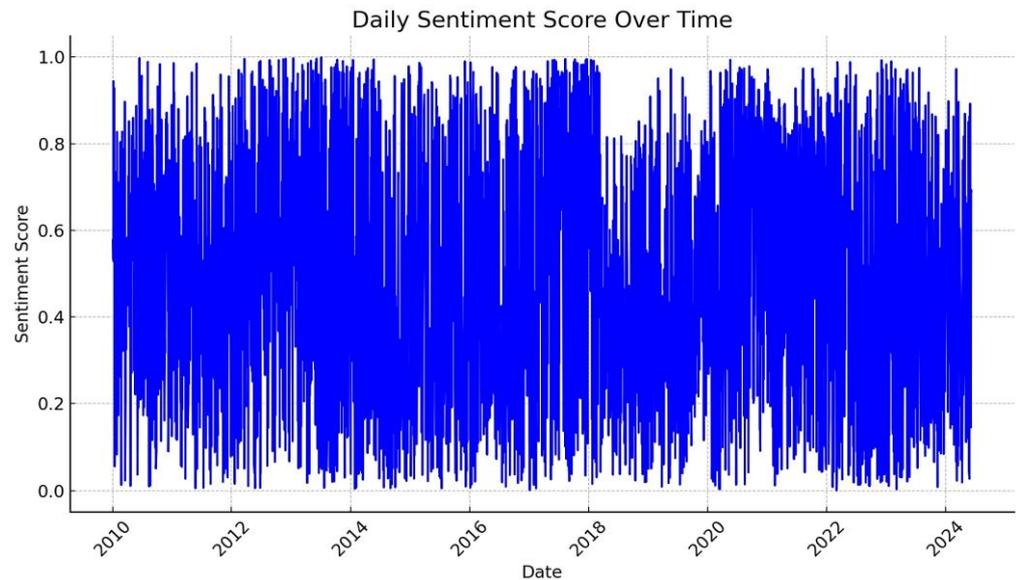

**Figure 7.** Sentiment score over time.

The NGX market's sentiment scores show significant volatility over time, with frequent fluctuations between 0 and 1. This indicates a high sensitivity to financial news and other exogenous variables. The repeated extreme fluctuations suggest a strong reaction to news. Certain time frames between 2012–2014 and 2017–2018 show more positive sentiment score clustering towards the upper end of the scale (closer to 1), while other periods show more balanced or lower sentiment scores. This could indicate more favorable market conditions or more pessimistic conditions in others (such as the lower sentiment score seen in 2023). The observable decrease in overall sentiment with lower clustering in 2023 reflects less optimistic market news, heightened volatility, and negative financial developments. Understanding these fluctuations is necessary for traders and investors, as sentiment-based trading algorithms can capitalize on the changes. Nevertheless, models must also be resilient to unforeseen sentiment fluctuations.

The histogram of predicted probabilities in Figure 8 shows an even distribution and a wide spread across the 0 to 1 range. This indicates uncertainty in market sentiment. It also suggests the market is often in flux, reacts to various factors, and often predicts price movements with caution or moderate confidence. The sentiment near 0.5 suggests the potential for either direction depending on subsequent financial news. This simply suggests adding more nuanced strategies like technical indicators and historical trends to account for market uncertainty rather than relying solely on news sentiment. The model makes relatively few predictions with probabilities below 0.30 or above 0.60. The lack of extreme probabilities implies that the model is cautious.

The findings of Bagate et al. on the application of sentiment analysis in algorithmic trading using various machine learning models, such as Logistic Regression, to predict stock market prices showed that Logistic Regression is prone to error risk but can be effective in the initial monitoring of price movement alone [19]. This aligns with the study findings, as sole reliance on sentiment for market direction could lead to missed or incorrect predictions. Instead, strategies should consider the inherent uncertainty and potential for varied market reactions. The sentiment score daily volatility suggests a reactive market environment, while the model's cautious probability distribution indicates that it is well calibrated to handle this volatility without making overly confident predictions. However, the model performs well but is not perfect. This shows that LR can distinguish between positive and negative outcomes with some uncertainty. Therefore, investment strategies should incorporate this uncertainty rather than relying solely on market sentiment.

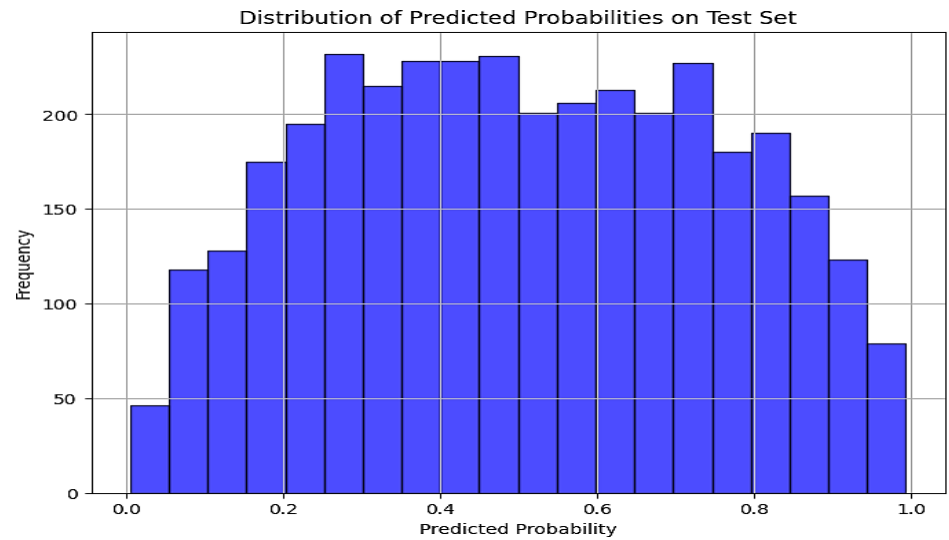

**Figure 8.** Distribution of predicted probabilities on test set.

## 4. Discussion

Table 4 shows the classification results of the sentiment analysis on financial news.

**Table 4.** Comparative analysis of sentiment analysis models.

| Test Set Metrics | GPT Predefined Approach (%) | FinBERT (%) | Logistic Regression (%) |
|---|---|---|---|
| Accuracy | 54.19 | 63.33 | 81.83 |
| Precision | 72.66 | 63.76 | 82.57 |
| Test Recall | 45.09 | 63.33 | 81.15 |
| Test F1 Score | 32.69 | 63.30 | 81.85 |
| Test ROC AUC | 65.37 | 65.59 | 89.76 |

The GPT predefined approach has high precision (72.66%), moderate ROC AUC (65.37%), and exhibits weakness with low recall (32.69%) and F1 score (45.06%). FinBERT has a balanced performance with a slightly better result than the GPT predefined approach but with lower precision (63.66%). FinBERT requires a large dataset and substantial computational resources and experience to fine-tune and implement. This is unlike Logistic Regression, which shows high accuracy (81.83%) and precision (82.57%) to reflect a strong overall performance. The model's F1 Score (81.85%) and ROC AUC (89.76%) are the highest, which indicates a good balance between precision and recall. Logistic Regression is simple and interpretable but may struggle with complex, non-linear relationships in the data.

The discrepancy in the sentiment distribution of the GPT predefined approach could be attributed to rules of simple assessments or heuristic methods associated with the predefined approach. The method often lacks flexibility and adaptability to new data patterns. It lacks the capacity for improvement and learning of a machine learning model, which can learn and provide more accurate sentiment predictions. Although the precision metric is high, the overall performance metrics show that the predefined approach may not generalize well to unseen data or may inherit bias from training data. The operation like black boxes by GPT may lead to questioning the result generated. Paripati et al. highlighted ethical issues, such as privacy and data protection, accountability and governance, bias and fairness, etc., when utilizing GPT models for data analysis [20]. Additionally, the performance of GPT-4, like any large language model, depends on the quality and diversity of its training data. In this study, the predefined GPT-4 approach was trained and tested using a large and varied set of financial news, with a total of 24,923 news headlines. This reduces the risk of overfitting, but the complexity of GPT-4 might not necessarily offer an advantage for a classification task where a simpler model, like Logistic Regression, could perform better. Additionally, GPT-4 might not fully grasp subtle financial terminology or the nuanced impact of economic terms on sentiment, especially when faced with new, specific contexts within the test data. This does not undermine GPT-4's capabilities but rather highlights the importance of context-specific tuning when dealing with highly specialized tasks. Additionally, the predefined sentiment approach might lack the flexibility to fully adapt to financial jargon and context-specific sentiment nuances, which could contribute to underperformance compared to models like Logistic Regression, which can be optimized for specific datasets [21]. In summary, the predefined GPT approach is a broad, generalized model that may lack the necessary domain-specific adaptations to handle the unique challenges of financial news sentiment classification. In contrast, Logistic Regression benefits from simplicity, interpretability, and the ability to be fine-tuned for the dataset, making it a better performer in this specific context.

Yang et al. demonstrated how FinBERT outperformed conventional models in capturing market sentiment in the sentiment analysis of financial texts [22]. This aligns with the study that confirmed FinBERT's ability to process unstructured text and extract insights for financial prediction and analysis. Although FinBERT had a balanced performance in this study, its use of complex embeddings and attention mechanisms did not necessarily translate to better performance compared to Logistic Regression on this specific dataset, where the sentiment signals were likely more straightforward. However, the complexity and high computational power and processing time of FinBERT make it a barrier to practical deployment. Logistic Regression, having the best result out of the three models examined, works well with structured data, can capture linear relationships, and is favored for transparent models. Although it has limitations with non-linear data, its ability to perform well on the financial dataset may likely be due to effective feature engineering and regularization. The simplicity of Logistic Regression compared to deep learning models like FinBERT and GPT-4 has a distinct advantage when applied to the well-defined financial news datasets used in this study that tend to contain regular patterns and structured sentiment indicators. In contrast, FinBERT and GPT-4 are designed for more complex tasks, which require substantial computational power and fine-tuning [23]. The strong performance of Logistic Regression could be attributed to the text dataset's features, which are well suited to linear separability. The model's simplicity ensures robustness and reduces the risk of overfitting. Additionally, effective feature engineering, such as using TF-IDF for text representation, further enhances Logistic Regression performance by making it focus on meaningful data points as well as providing structured, high-quality inputs that improve the model's efficiency [24]. Additionally, simple models like Logistic Regression can outperform and be more effective than complex models when clear and structured signals (such as the presence of positive or negative words to describe market conditions) are used in the financial sentiment analysis. These models avoid unnecessary complexity and focus on key features extracted from well-prepared text data (like using TF-IDF). As a result, Logistic Regression tends to generalize better to new, unseen data. This approach aligns with Occam's Razor, a principle that suggests that simpler models are often better suited for tasks where complexity does not lead to a meaningful improvement in accuracy. Zhang L et al.'s (MIT researchers) study on financial forecasting models showed that in volatile market conditions, a simple model like Logistic Regression provided more reliable forecasts within shorter time frames, while neural networks and deep learning models tended to struggle with noisy data, thereby reinforcing the application of Occam's Razor in financial modeling [25]. The benefits of Logistic Regression from well-engineered input features and regularization techniques like L2 help prevent overfitting by penalizing large coefficients, and this further enhances the model's performance by improving generalization. Kumar and Elakkiya highlighted Logistic Regression effectiveness in financial risk prediction when combined with feature selection techniques [26]. The model performed well by leveraging key financial signals embedded in the text without the added complexity of needing to understand deeper context or sentiment nuances, as required by FinBERT or GPT-4. The ease of use and implementation of Logistic Regression are important in financial applications. Logistic Regression is the best choice for the news dataset due to its high accuracy, precision, and F1 Score. Additionally, Logistic Regression is the best-performing distribution in terms of predicted probabilities because it shows a more balanced distribution of predictions among the three models. It allows for some confident predictions (both low and high probabilities) while still maintaining caution in other cases. This balance is often desired in machine learning models, where both over-confidence and under-confidence can be detrimental. FinBERT and GPT distribution of predicted probabilities plots are overly cautious, and this may limit their practical usefulness in making real-world predictions. Furthermore, the daily sentiment score over time generated by Logistic Regression has high-level details and sensitivity to the smallest sentiment changes compared to other models. This is ideal for situations where sentiment volatility, extreme events, and high-frequency data are critical for stock market decision-making, such as in-depth statistical analysis or high-frequency trading systems. The study conducted by Huang et al. supported the notion of the study. The study is a comparative analysis of models used for sentiment analysis in financial markets, and their findings indicated that Logistic Regression offers high granularity and reacts more sensitively to minor fluctuations in sentiment data, which makes it more suitable for high-frequency trading scenarios [12]. The findings were evident when the model was applied to financial news datasets, where Logistic Regression outperformed more complex models in capturing rapid sentiment shifts. FinBERT, which provides a competitive metric and excellent sentiment analysis, may not be suitable for this dataset due to its complexity and resource demands. Although FinBERT provided competitive metric results compared to Logistic Regression, its complexity and resource demands may not warrant its utilization over Logistic Regression for this financial news dataset. The study did not rule out the consideration of using FinBERT for future exploration in specific tasks that involve nuanced sentiment analysis in financial texts.

Based on the findings of this study, we recommend prioritizing Logistic Regression for NGX index sentiment prediction tasks, especially when computational efficiency and model simplicity are important. Its high accuracy, combined with minimal computational resources, makes it a practical choice for real-time market forecasting. However, FinBERT and GPT-4 should not be overlooked. Their ability to analyze complex textual data and understand nuanced sentiment is valuable for comprehensive stock market sentiment prediction models. For future research, we recommend exploring hybrid models that combine the strengths of Logistic Regression's simplicity and accuracy with the depth of sentiment analysis provided by FinBERT and GPT-4. Furthermore, additional external data sources, such as macroeconomic indicators or geopolitical news, should be integrated into these models to improve predictive accuracy. This would allow for a more holistic view of the factors influencing stock market behavior, which would lead to better forecasts. Finally, the continued use of hyperparameter optimization tools, such as Optuna, is crucial for ensuring the models perform at their best across different datasets and market conditions.

## 5. Conclusions

This study provides a comprehensive analysis of the predictive capabilities of FinBERT, GPT-4, and Logistic Regression for stock index prediction using the NGX All-Share Index dataset categorization. The results indicate that Logistic Regression, despite being a simpler model, outperformed FinBERT and GPT-4 in terms of accuracy, precision, recall, F1 score, and ROC AUC. The robustness of Logistic Regression, particularly after hyperparameter tuning, made it the most efficient model for predicting stock market trends, achieving a high accuracy of 81.83% and ROC AUC of 89.76%. The superior performance of Logistic Regression can be attributed to several factors, which include its ability to handle structured data efficiently, reduced risk of overfitting compared to more complex models, the success of feature engineering and regularization, and its inherent simplicity and interpretability. FinBERT, while better equipped to handle financial language, faced challenges in terms of computational demands and resource usage, which limited its practical application in real-time prediction scenarios. Although GPT-4 is powerful in general text analysis, it showed limitations when applied specifically to financial data.

These findings suggest that, while advanced NLP models offer promise in sentiment analysis, traditional models like Logistic Regression still provide strong performance with lower computational costs. The study also highlights the importance of selecting the right model for the task at hand, and in this case, the simplicity and effectiveness of Logistic Regression proved to be the best fit. However, FinBERT and GPT-4 offer avenues for future exploration, especially when combined with other machine learning techniques in a hybrid approach.

*Policy Recommendations*

- Hybrid Model Approaches: FinBERT and GPT-4, despite their current limitations in this study, should not be dismissed in financial prediction tasks. Future research should focus on integrating these advanced NLP models with traditional machine learning models, such as Logistic Regression or other ensemble methods, to leverage the strengths of both approaches in emerging markets. For instance, FinBERT can be used for sentiment analysis in combination with Logistic Regression for numerical prediction tasks, potentially improving overall model performance.
- Computational Efficiency Optimization: To address FinBERT's high computational demands, policy-makers and researchers should invest in improving the efficiency of such models. This may involve the development of more resource-efficient variants of FinBERT or the use of transfer learning techniques that reduce the need for extensive computational power without compromising accuracy. Public and private investment in accessible, high-performance computing infrastructure could also lower the barriers to using these models in real-time applications.
- Sector-Specific Fine-Tuning: The findings highlight the general limitations of the predefined approach of GPT-4 when applied to financial data. Future studies should explore fine-tuning GPT-4 (and similar models) specifically for financial forecasting tasks. This involves adapting the model to financial language, industry-specific terminology, and numerical prediction, which could significantly improve its predictive capability. Policy-makers in financial institutions should also encourage partnerships between academic researchers and AI practitioners to develop sector-specific NLP solutions.
- Practical Applications in Financial Markets: Financial institutions and regulators should be cautious when adopting advanced NLP models for real-time stock prediction due to their computational demands and current limitations in specialized financial analysis. However, these models may serve as valuable tools in combination with traditional methods for in-depth analysis, market sentiment assessment, or risk management. Regulatory bodies might consider establishing guidelines that encourage the responsible adoption of these technologies, ensuring a balance between innovative AI applications and financial stability.

In conclusion, while traditional models such as Logistic Regression currently offer strong performance at a lower computational cost, future research should not overlook the potential of FinBERT and GPT-4. The development of hybrid models, optimization of computational efficiency, and fine-tuning of models for specific domains will be critical in advancing AI-driven financial prediction.

**Author Contributions:** Conceptualization, S.A.-L. and O.S.; methodology, S.A.-L. and O.S.; software, S.A.-L. and O.S.; validation, O.S., O.P. and B.O.; formal analysis, S.A.-L. and O.S.; investigation, S.A.-L.; resources, O.S.; data curation, S.A.-L.; writing—original draft preparation, S.A.-L.; writing—review and editing, O.S., O.P. and B.O.; supervision, O.S.; project administration, O.P., B.O. and O.S. All authors have read and agreed to the published version of the manuscript.

**Funding:** This research received no external funding.

**Institutional Review Board Statement:** Not applicable.

**Informed Consent Statement:** Not applicable.

**Data Availability Statement:** Data are available upon request.

**Conflicts of Interest:** The authors declare no conflicts of interest.